\documentclass[]{article}

\usepackage{microtype}
\usepackage{graphicx}
\usepackage{textcomp}
\usepackage{subfig}
\usepackage[detect-all]{siunitx}
\usepackage{booktabs} 

\usepackage{hyperref}

\usepackage{dirtytalk}
\usepackage[accepted]{icml2022}

\usepackage{amsmath}
\usepackage{amssymb}
\usepackage{mathtools}
\usepackage{amsthm}

\usepackage[capitalize,noabbrev]{cleveref}

\usepackage[textsize=tiny]{todonotes}

\icmltitlerunning{Causal Discovery using Model Invariance through Knockoff Interventions}

\begin{document}

\twocolumn[
\icmltitle{Causal Discovery using Model Invariance through Knockoff Interventions}



\icmlsetsymbol{equal}{*}

\begin{icmlauthorlist}
 \icmlauthor{Wasim Ahmad} {1}
 \icmlauthor{Maha Shadaydeh} {1}
 \icmlauthor{Joachim Denzler} {1,2}
 \end{icmlauthorlist}

\icmlaffiliation{1}{Computer Vision Group, Freidrich Schiller University, Jena, Germany}
\icmlaffiliation{2}{Michael Stifel Center Jena, Germany}

\icmlcorrespondingauthor{Wasim Ahmad}{wasim.ahmad@uni-jena.de}
\icmlkeywords{Causal Inference, Model Invariance, Nonlinear Time series, Interventions}

\vskip 0.3in
]



\printAffiliationsAndNotice{}  

\begin{abstract}
Cause-effect analysis is crucial to understand the underlying mechanism of a system. We propose to exploit model invariance through interventions on the predictors to infer causality in nonlinear multivariate systems of time series. We model nonlinear interactions in time series using DeepAR and then expose the model to different environments using Knockoffs-based interventions to test model invariance. Knockoff samples are pairwise exchangeable, in-distribution and statistically null variables generated without knowing the response. We test model invariance where we show that the distribution of the response residual does not change significantly upon interventions on non-causal predictors. We evaluate our method on real and synthetically generated time series. Overall our method outperforms other widely used causality methods, i.e,  VAR Granger causality, VARLiNGAM and PCMCI$^+$. The code and data can be found at: \href{https://github.com/wasimahmadpk/deepCausality.git}{https://github.com/wasimahmadpk/deepCausality}
\end{abstract}

\section{Introduction}
\label{Introduction}

Discovering causal graph in a system is vital to understand system\textquotesingle s dynamics and its response to changes in environment. The methods which extract causal information from observational data are crucial for the broad field of artificial intelligence. In this work we use deep networks to learn nonlinear interactions in multivariate time series and test model invariance through Knockoffs \cite{Barber_2015, candes2017panning} based interventions for causal inference. We build our work on the assumption that the causal mechanism in a real-world system does not change across diverse environments \cite{peters2016causal, heinze2018invariant, pfister2019invariant}. More specifically, the conditional distribution of the response given its causal predictors remains invariant for all observations over different settings. A model is regarded as invariant if the distribution of the residuals of model\textquotesingle s response does not change when intervened on its non-causal predictors.

Deep networks best suit to learn nonlinear interactions in a multivariate system. We use DeepAR \cite{salinas2020deepar} to model nonlinear relations in multivariate time series, which is based on recurrent neural networks (RNNs) and generate probabilistic forecast. It characterizes conditioning on observed variables and estimate latent features i.e. trends and seasonality from time series that might help in the identification of spurious associations. To expose trained models to unseen settings for causal inference, we can either omit or replace model predictors with a different representation. Since, deep networks cannot handle missing input or out-of-distribution (OOD) data, we propose to use the theoretically well-established Knockoffs framework to generate interventional environments. 

Knockoffs are statistically null-variables, in-distribution and as uncorrelated as possible with the originals. Moreover, Knockoffs possess the pairwise exchangeability property when means that replacing any variable with their knockoff copy in a multivariate setup will not change the underlying joint distribution of the set of variables. We highlight the effectiveness of Knockoffs as interventional variables by comparing it with other intervention methods such as OOD, mean and uniform distribution. We discover full causal graph assuming causal sufficiency and stationarity in synthetic and real data and show that our method outperforms widely used causality methods, i.e., VAR Granger causality (VAR-GC) \cite{granger1969investigating}, VARLiNGAM \cite{hyvarinen2010estimation}, PCMCI$^+$ \cite{runge2020discovering} as well as other applied intervention methods.

The remaining of this paper is organized as follows. In Section \ref{section:relatedwork}, we cover the related work. We provide methodological background in Section \ref{section:methods}. In Section \ref{section:causality}, we describe our proposed method. Results are presented and discussed in Section \ref{section:experiments}. In Section \ref{section:conclusion} we conclude our work.
\section{Related Work}
\label{section:relatedwork}

The methods in \cite{peters2016causal, heinze2018invariant, pfister2019invariant, gamella2020active, martinet2021variance} use invariant causal prediction (ICP) for causal inference assuming that data samples are drawn from heterogeneous environments and can identify the ancestors of the response. These methods are limited in a sense that they estimate causal relationship of the candidate variable with the target only rather than discovering full causal graph as done in our work through multivariate target modelling. Moreover, they search for subset of predictors which remains unchanged throughout environments while we estimate a consistent causal set using model invariance across interventional environments over multiple forecast windows. ICP outputs the intersection of all subsets of invariant predictor which forms a strict subset of all direct causes or may even be empty. This is because disjoint sets of predictors can be invariant, yielding an empty intersection. In \cite{mogensen2022invariant}, the authors introduce and characterize minimally invariant sets of predictors, that is, invariant sets $S$ for which no proper subset is invariant. They propose to consider the union S$_{\text{IAS}}$ of all minimally invariant sets, where IAS stands for invariant ancestry search.

The method in \cite{peters2016causal} works with non-stationary data assuming that the environments are known in order to exploit the invariance. It is argued that without knowledge of the environments, the causal inference becomes difficult. Moreover, estimating the environments from data in naive manner e.g.,  clustering, change point detection, etc., and subsequently applying the existing methodology may lead to a loss of the methods coverage, guarantee or yield less powerful results. The authors of \cite{pfister2019invariant} do not estimate the environments but instead utilizes the existing non-invariances in observations. In our method, we generate knockoff representation of the predictors to represent different environments and test model invariance for causal inference in nonlinear systems. Previously we used deep networks with Knockoffs-based interventions to estimate Granger causality in nonlinear multivariate time series \cite{ahmad2021causal}. 

The authors of \cite{Tank_2021, romano2020deep} utilize deep networks combined with sparsity-inducing penalties on the weights to estimate causal significance of the predictors. However, it is difficult to retrieve clear information from weights of the deep networks that can be used for interpretability \cite{luo2020causal}. The widely used PCMCI \cite{runge2019detecting} method applies linear and nonlinear conditional independent (CI) tests to infer causality in multivariate time series. The extended version of cause-effect variational autoencoders (CEVAE) \cite{louizos2017causal} integrates the domain-specific knowledge for nonlinear causal inference in ecological time series \cite{trifunov2019nonlinear}. Time-frequency analysis has been done to handle spurious associations in time series \cite{shadaydeh2019time} however, they use spectral representation of linear VAR-GC and might struggle in case of nonlinearity.
\section{Methodological Background}
\label{section:methods}
In this section we introduce the necessary background on DeepAR and the Knockoffs framework. 

\subsection{DeepAR}
DeepAR is a powerful method suitable for probabilistic forecasting of high dimensional nonlinear non-stationary time series \cite{salinas2020deepar}. It uses LSTM-based autoregressive RNNs \cite{graves2013generating} and learns nonlinear interactions from the history of all the related time-series. Few key attributes of DeepAR are: It extracts hidden features in the form of trends and seasonality in data, and characterizes conditioning on the entire set of variables during inference which can be helpful in identifying spurious relations. Moreover, lesser hand-crafted feature engineering is needed in DeepAR to capture complex, group-dependent behavior.

\subsection{Knockoffs} 
The Knockoff framework is developed to estimate feature importance using CI testing with controllable false discovery rate \cite{Barber_2015}. Given a set of observed variables $Z = (Z_{1}, \dots, Z_{n})$ in an environment with known distribution $P_Z$, and a predictive model, the knockoff copies of the observed variables, defined as $\widetilde{Z} = (\widetilde{Z}_{1}, \dots, \widetilde{Z}_{n})$, represent interventional environments while maintaining the in-distribution and decorrelation property of the observational environment. Knockoff samples are statistically null-variables, generated without knowing model response. Moreover, knockoffs should satisfy the pairwise exchangeability condition \cite{Barber_2015}
\begin{equation}
\label{eqn:exchange}
(Z_1, {\widetilde{Z}_2}, \widetilde{Z}_1, {Z_2}) \overset{d}{=} (Z_1, Z_2, \widetilde{Z}_1, \widetilde{Z}_2)
\end{equation}
for any subset $ \mathbb{A} \subseteq {1,\dots,n}$, here $\overset{d}{=}$ represents equal distributions. 
The $(Z, \widetilde{Z})_{swap(\mathbb{A})}$ is obtained from $(Z, \widetilde{Z})$ by swapping the entries $Z_j$ and $\widetilde{Z}_j$ for each  $j \in \mathbb{A}$. 

The knockoff mechanism can be thought of as generating a probability distribution $P_{\widetilde{Z}|Z} (.|z)$ which is the conditional distribution of $\widetilde{Z}$ given $Z=z$ chosen such that the obtained joint distribution of $(Z, \widetilde{Z})$ being equal to 

\begin{center}
    $P_{Z}(z) P_{\widetilde{Z}|Z}(\widetilde{z}|z)$,
\end{center} 

is pairwise $(z_j, \widetilde{z}_j)$ symmetric and satisfy the exchangeability condition in \eqref{eqn:exchange}. For variables with Gaussian distribution $P_Z = \mathcal{N}_n(\textbf{0}_n, \Sigma)$,  \cite{candes2017panning} shows that knockoffs $\widetilde{\textbf{Z}}_{i, *}$ can be drawn from the conditional distribution
\begin{equation}
\label{eqn:condist}
P_{\widetilde{Z}|Z}(.|\textbf{Z}_{i, *}) = \mathcal{N}_n((\textbf{I}_n - S\Sigma^{-1})\textbf{Z}_{i, *}, 2S -S \Sigma^{-1} S)
\end{equation} 
for any fixed diagonal matrix $S$ satisfying $0\leq S \leq 2\Sigma$. This leads to joint distribution of $(\textbf{Z}_{i, *}, \widetilde{\textbf{Z}}_{i, *})$ being  equal to
\[\mathcal{N}_{2n}
\begin{pmatrix}
\textbf{0}_{2n},
\begin{bmatrix}
\Sigma & \Sigma -S\\
\Sigma - S & \Sigma
\end{bmatrix}
\end{pmatrix},\] which satisfies the pairwise exchangeablity \eqref{eqn:exchange}. Here $S$ is to be maximized which represents the difference between the original variables and their knockoff version. As a result the off-diagonal term $\Sigma - S$ is minimized making the covariance of the original variable with its corresponding knockoff version as low as possible.
\section{Causality using Model Invariance through Knockoffs}
\label{section:causality}

Let $z_i, i = 1,\dots,  N$ be the $N-$variate time series. Each time series $z_{i, t}, t = 1, \dots, r$ is a realization of length $r$ real-valued discrete stochastic process $Z_i, i = 1,\dots, N$. Our goal is to estimate the causal relations among all variables and retrieve a full causal graph. Throughout our work, we assume stationarity and causal sufficiency which implies $Z_i, i = 1,\dots, N$  is a stationary stochastic process, and the set of observed variables $Z_i, i = 1, \dots, N$  includes all of the common causes of pairs in $Z$. An extension of the current work to deal with non-stationary settings is in progress. 

We exploit the idea of model invariance for causal discovery in nonlinear time series where we determine that the underlying causal structure of a system remains invariant over interventional environments.  We model nonlinear time series with DeepAR and generate knockoff copies of the original data to test model invariance. A model is regarded invariant if the distribution,  i.e.,  $\mu$ and $\sigma^2$,  of the response residuals given its causal predictors do not change against interventional environments. The predictors are considered as non-causal if the residual distribution of the model response does not significantly change by exposing it to their knockoff representations. We retrieve causal graph with a consistent set of variables using hypothesis testing over residual distribution from multiple forecast windows.

 
%

To generate knockoff variables, we implement DeepKnockoffs \cite{romano2020deep}, a framework for sampling approximate knockoffs. We use mixture of Gaussian models as implemented in \cite{gimenez2019knockoffs} to obtain knockoff samples. The procedure consists of first sampling the mixture assignment variable from the posterior distribution. The knockoff variables are then sampled from the conditional distribution of the Knockoffs given the original variables and the sampled mixture assignment such that the necessary conditions for Knockoffs are satisfied. Swapping any knockoff variable with their original version must leave the underlying joint data distribution invariant as in \eqref{eqn:exchange}. Given observed variable $Z \sim \mathcal{N}(\mu, \Sigma)$, we aim to build its knockoff copy $\widetilde{Z}$ in such a way that the joint distribution of $(Z, \widetilde{Z}) \sim \mathcal{N(*, **)}$ is Gaussian and 
\[
* =
\begin{bmatrix}
\mu \\
\mu
\end{bmatrix}, 
** =
\begin{bmatrix}
\Sigma & \Sigma -S\\
\Sigma - S & \Sigma
\end{bmatrix},\]
where $\textit{S}$ is a diagonal matrix such that $** \succeq 0$. $\textit{S}$ represents the discrepancy between originals and knockoffs. We have the same $\mu$ for both original and knockoff version. The Knockoff framework maintain the same covariance among variables in set of generated Knockoffs as in originals which is represented by the diagonal entries in given covariance matrix. However, the covariance $\Sigma - S$ of the original variable with its corresponding knockoff copy in the off-diagonals is kept as low as possible by maximizing $\textit{S}$. We obtain Knockoffs $\widetilde{Z}_{j}, j \subseteq {1, \cdots,n}$, for all variables in the multivariate time series and substitute one variable $Z_j$ at a time with $\widetilde{Z}_{j}$ for each $j \subseteq {1, \cdots,n}$ to test model invariance and estimate cause-effect relationships.

\begin{figure*}
    \centering
    \subfloat[\centering Causal case]{{\includegraphics[width=7.00cm]{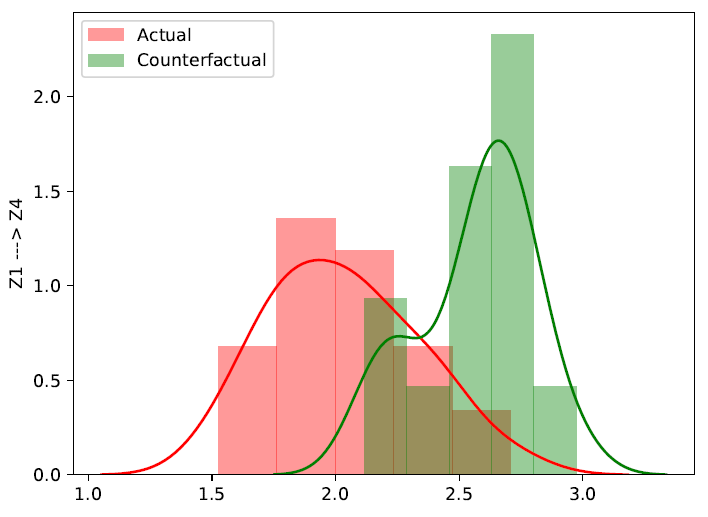} }}
    \qquad
    \subfloat[\centering Non-causal case]{{\includegraphics[width=7.00cm]{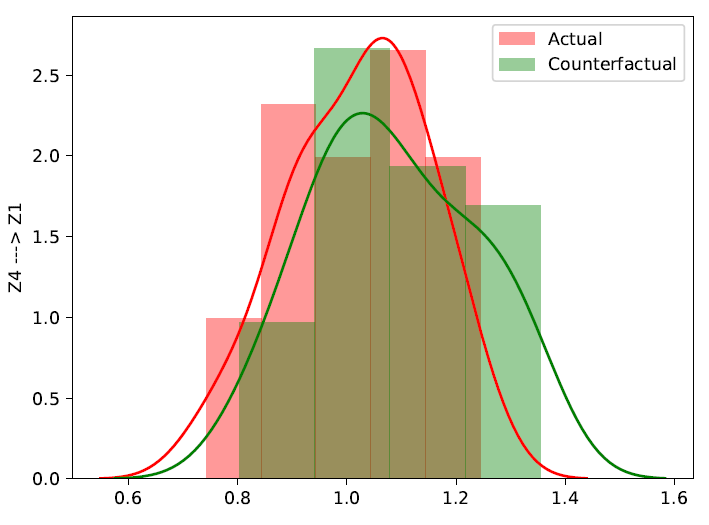} }}
    \caption{(a) Model residual distribution for causal scenario where $Z_1$ causes $Z_4$ in our generated causal graph. (b) Model residual distribution for non-causal scenario where $Z_4$ is not causal for $Z_1$ based on ground truth topology.}
    \label{fig:residuals}
\end{figure*}
To estimate causal effect of $z_i,i = 1,\dots, N$ on the response $z_j$, we test model invariance using its forecast residuals by feeding a knockoff representation $\widetilde{z}_i$ of the original variable $z_i$. We obtain residuals $e_1, e_2, ..., e_n \sim R$ and $\Tilde{e}_1, \Tilde{e}_1, ..., \Tilde{e}_n \sim \Tilde{R}$ over forecast windows $w$ in range $[20 - 30]$ with a step size of $[5 - 10]$ for each realization $z_{j,t}$ of the  stochastic process $Z_j, j = 1, \dots, N$ with and without intervention on $z_{i}$, $i \neq j$. $R$ represents distribution of model residuals for actual outcome and $\Tilde{R}$ shows residual distribution after intervention. We compare these two distributions using Kolmogorov–Smirnov (KS) test \cite{smirnov1939estimation} for causal inference. The test statistic uses the supremum distance $D_{n,m}$ between $R$ and $\Tilde{R}$. Here $n, m$ are the number of samples in the these two distributions.
\begin{equation}
\label{eqn:kstest}
D_{n, m} = \sqrt{\frac{nm}{n + m}}sup|R_n - \Tilde{R}_m |
\end{equation}
The  null hypothesis $H_0$:  $z_i$ does not cause $z_j$, is accepted if the distribution of the residuals is approximately identical across environments $\Tilde{R} \overset{d}{\approx} R_i$ with a significant level $\alpha$ in range $[5 - 10]\%$ . In case the residual distribution is significantly different $\Tilde{R} \overset{d}{\neq} R_i$, the alternate hypothesis $H_1$: $z_i$ causes  $z_j$ is accepted. $D_{n,m}$ tends to have significantly larger value in alternate case.

\section{Experimental data and Results}
\label{section:experiments}
We conducted experiments on real and synthetically generated time series data. The synthetic data model has the generic form:
\begin{equation}
\label{eqn:syndata}
Z_t^j = a_{j}Z_{t-1}^j + \Sigma_i c_i f_i (Z_{t-\tau_i}^i) + \eta_t^j
\end{equation}
for $j \in {1, ..., N}$. The correlation coefficients $a_j$ are selected from range $[0.2 - 1.0]$. Besides autodependency in the model, there exist linear and exponential functional dependencies $f_{i}(z)$. We included noise $\eta_t^j$, which represents data from uncorrelated, normally distributed noise with zero mean and variances in the range $[0.30 - 0.9]$. The coupling coefficients $c_i$ and lags $\tau_i$ between variables are set to $[0.2 - 1.0]$ and $[0 - 10]$ respectively. To test our method, we generate multiple causal graphs with $N=5$ nodes by changing functional dependencies, causal edges, time delays, dynamic noise and  correlation coefficients throughout our experiments. To quantify the performance of applied methods, we calculate false positive rate (FPR) and F-score metrics in a variety of generated causal graphs.


To show the effectiveness of proposed approach where we utilize Knockoffs-based intervention variables, we compare its performance against other widely used causal inference methods. Besides that, to highlight the power of using Knockoffs as intervention variables, we test the proposed approach with the following methods for generating intervention variables. 

Distribution mean: The predictor $z_i$ is replaced with $\overline{z}_i$, that is comprised of the mean of the original variable and Gaussian noise.  $\overline{z}_i=\frac{1}{r}\sum_{t=1}^{r} z_{i, t} + \mathcal{N}(0, \Sigma)$. 

Uniform distribution: The original variable $z_i$ is replaced by a variable $z'_i$ having a uniform distribution ranging from minimum to maximum values of the  original variables $z_i$  as suggested in \cite{jiwoong2021causal}. 

Out-of-distribution: The variable $z_i$ having a distribution $D_i$ is replaced with another variable sampled from OOD $\overline{D}$ such that $\overline{D} \overset{d}{\neq} D_i$, different $\mu$ and $\sigma^2$. Moreover, it is generated to be uncorrelated with the original variables.

As a real-world application, we conducted experiments on average daily discharges of rivers in the upper Danube basin, provided by the Bavarian Environmental Agency \footnote{https://www.gkd.bayern.de}. We use measurements from 2017 to 2019 from the Iller at Kempten $K_t$, the Danube at Dillingen $D_t$, and the Isar at Lenggries $L_t$. The Iller flows into the Danube within a day, which implies an instantaneous causal link $K_t \rightarrow D_t$ and no direct causal links between the pairs $K_t$, $L_t$ and $D_t$, $L_t$. Moreover all variables may be confounded by rainfall or other weather conditions \cite{gerhardus2020high}, which test the ability of the applied methods to distinguish between spurious associations and true causal relationships.

\begin{figure*}
    \centering
    \subfloat[\centering False Positive Rate]{{\includegraphics[width=8.1cm]{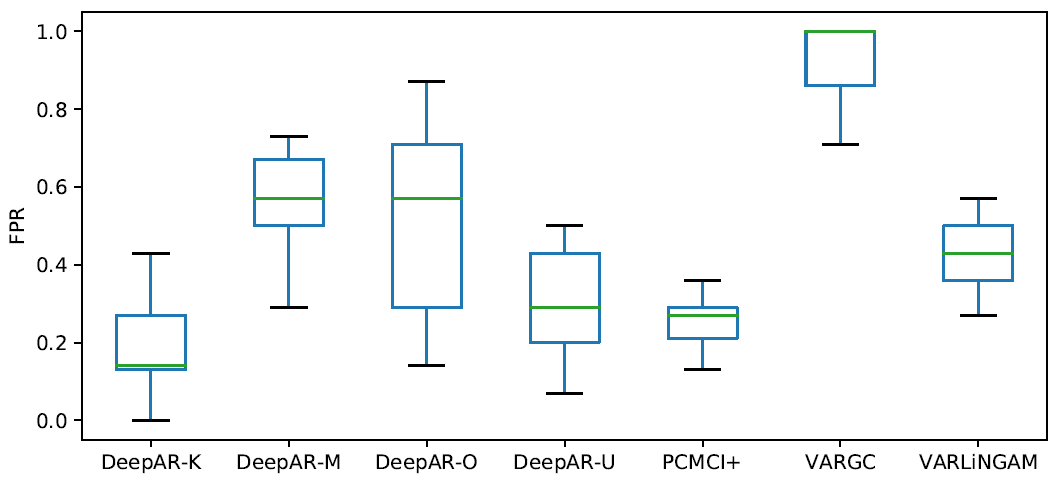} }}
     \qquad
    \subfloat[\centering F-score]{\includegraphics[width=8.1cm]{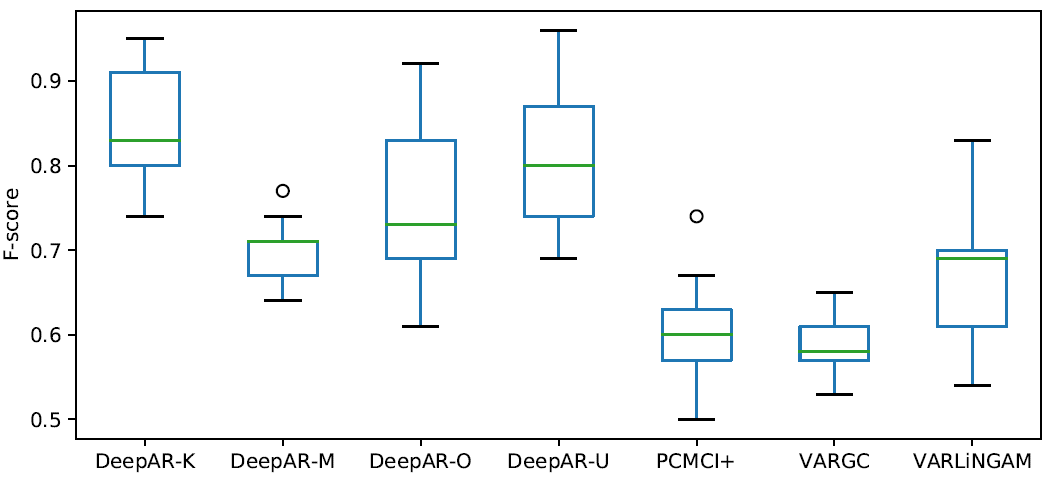}}
    \caption{(a) False positive rate (FPR) of used methods for causal inference in synthetic multivariate nonlinear time series, lower is better. (b) F-score of all applied methods for estimating causality in synthetic multivariate nonlinear time series, higher is better.}
    \label{fig:results}
\end{figure*}
\begin{table*}[ht]
\small
\centering
\caption{Causality in river discharges time series data by VAR-GC, VARLiNGAM, PCMCI$^+$ and DeepAR with applied interventions types. The detection of a causal link between variables is represented by $\checkmark$ and absence of causality is shown by $\times$.}
\begin{center}
\begin{tabular}{p{1.50cm}p{1.25cm}p{1.25cm}p{1.4cm}p{1.25cm}p{1.45cm}p{1.35cm}p{1.35cm}p{1.35cm}}
    \hline
    Pairs & Expected links & VAR-GC & VAR-LiNGAM & PCMCI$^+$ & DeepAR-Knockoffs & DeepAR-OOD & DeepAR-Mean & DeepAR-Uniform\\
    \hline
    $K_t \rightarrow D_t$ & \checkmark & \checkmark & \checkmark & \checkmark & \checkmark & \checkmark & $\times$  & \checkmark \\

    $K_t \rightarrow L_t$ & $\times$ & $\times$ & \checkmark & $\times$ & $\times$ & $\times$ & $\times$ & $\times$ \\

    $D_t \rightarrow K_t$ & $\times$ & $\times$ & $\times$ & $\times$ & $\times$ & \checkmark & $\times$ & $\times$ \\

    $D_t \rightarrow L_t$
    & $\times$ & \checkmark & $\times$ & $\times$ & $\times$  & $\times$ & \checkmark & \checkmark\\

    $L_t \rightarrow K_t$ & $\times$ 
    & \checkmark & \checkmark & \checkmark & $\times$  & $\times$ & \checkmark & $\times$\\

    $L_t \rightarrow D_t$
    & $\times$ & \checkmark & $\times$ & $\times$ & $\times$  & $\times$ & \checkmark & $\times$\\
    \hline
\end{tabular}
\end{center}
\label{tab:river1}
\end{table*}

\textbf{Results}: We generate multiple causal graphs using synthetic data model and demonstrate the results in terms of FPR and F-score for VAR-GC, VARLiNGAM, PCMCI$^+$ and DeepAR with all four types of intervention variables represented as DeepAR-K for Knockoffs, DeepAR-O for OOD, DeepAR-M for mean and DeepAR-U for uniform distribution. We infer causality among nodes of the graph using model invariance through knockoff interventions where we use the KS test (\ref{eqn:kstest}) to compare the distribution of model residuals with and without interventions. In Fig \ref{fig:residuals}, we illustrate the causal decision by our method between pair of nodes $Z_1$ and $Z_4$. Based on the ground truth from our generated causal graph, $Z_1$ causes $Z_4$ while the reverse in not true. Our method accurately identifies $Z_1 \rightarrow Z_4$, a true positive and $Z_4 \nrightarrow Z_1$, a true negative. Overall DeepAR-K performs better than others in accurate causal discovery. Since the Knockoffs-framework is originally developed  to control false discoveries, we can see low FPR for Knockoffs-based interventions as shown in Fig \ref{fig:results} (a) compared to other methods and intervention types. At the same time, the power of DeepAR-K to identify true causal links is high as illustrated in Fig \ref{fig:results} (b),  which could be attributed to the uncorrelated property of the Knockoffs with their originals. The closest to our proposed DeepAR-K in terms of achieving high F-score and low FPR is DeepAR-U. PCMCI$^+$ also achieves relatively better FPR but shows low power for true causal detection along with VAR-GC and VARLiNGAM. The low detection power of these methods might be because of nonlinearity in the generated causal graphs.








For river discharges data, we have one contemporaneous link $K_t \rightarrow{D_t}$ as per description in the dataset, and we expect the applied methods to identify it. Our proposed method along with all other methods except DeepAR-Mean accurately detected $K_t \rightarrow{D_t}$ as shown in Table \ref{tab:river1}. Moreover, our method does not report any false causal link. However, the rest of the methods show spurious associations between various pairs of river discharges time series. DeepAR-OOD incorrectly detected the reversed causal link $D_t \rightarrow{K_t}$ too. VAR-GC and DeepAR-M identified a bidirectional link between $D_t$ and $L_t$ which may be because of the weather acting as a confounder since there is no expected direct causal relationship. DeepAR-U, DeepAR-OOD and PCMCI$^+$ have one false link detection each in different pairs.

\section{Conclusion}
\label{section:conclusion}
We proposed to test  model invariance across interventional environments over multiple forecast windows to discover full causal graph in multivariate nonlinear systems. DeepAR is used to learn nonlinear relationship in multivariate time series. The model invariance is tested via Knockoffs-based interventions for causal inference. We used synthetic and real time series data to evaluate applied methods. The results indicate that the estimation of the environments using Knockoff framework to test model invariance for causal inference yield better results in comparison to other intervention types. Results also show that the proposed method outperforms VAR-GC, VARLiNGAM and PCMCI$^+$, however having a higher computation time. Further we aim to extend the current work for non-stationary settings and test our method on other real-world time series.

\section*{Acknowledgements} 
This work is funded by the Carl Zeiss Foundation within the scope of the program line \say{Breakthroughs: Exploring Intelligent Systems} for \say{Digitization — explore the basics, use applications} and the DFG grant SH 1682/1-1.

\bibliography{example_paper}
\bibliographystyle{icml2022}

\newpage


\end{document}